# A novel georeferenced dataset for stereo visual odometry

Ivan Krešo
University of Zagreb Faculty of
Electrical Engineering and Computing
Zagreb, HR-10000, Croatia
Email: ivan.kreso@unizg.fer.hr

Marko Ševrović
University of Zagreb Faculty of
Transport and Traffic Engineering
Zagreb, HR-10000, Croatia
Email: sevrovic@unizg.fpz.hr

Siniša Šegvić
University of Zagreb Faculty of
Electrical Engineering and Computing
Zagreb, HR-10000, Croatia
Email: sinisa.segvic@unizg.fer.hr

*Abstract*—In this work, we present a novel dataset for assessing the accuracy of stereo visual odometry. The dataset has been acquired by a small-baseline stereo rig mounted on the top of a moving car. The groundtruth is supplied by a consumer grade GPS device without IMU. Synchronization and alignment between GPS readings and stereo frames are recovered after the acquisition. We show that the attained groundtruth accuracy allows to draw useful conclusions in practice. The presented experiments address influence of camera calibration, baseline distance and zero-disparity features to the achieved reconstruction performance.

## I. INTRODUCTION

Visual odometry [1] is a technique for estimating ego-motion [2] of a monocular or multiple camera system from a sequence of acquired images. The technique is interesting due to many interesting applications such as autonomous navigation or driver assistance, but also because it forms the basis for more involved approaches which rely on partial or full 3D reconstruction of the scene. The term was coined as such due to similarity with classic wheel odometry which is a widely used localization technique in robotics [3]. Both techniques estimate the current location by integrating incremental motion along the path, and are therefore subject to cumulative errors along the way. However, while the classic odometry relies on rotary wheel encoders, the visual odometry recovers incremental motion by employing correspondences between pairs of images acquired along the path. Thus the visual odometry is not affected by wheel slippage in uneven terrain or other poor terrain conditions. Additionally, its usage is not limited to wheeled vehicles. On the other hand, the visual odometry can not be used in environments lacking enough textured static objects such as in some tight indoor corridors and at sea. Visual odometry is especially important in places with poor coverage of GNSS (Global Navigation Satellite System). signal, such as in tunnels, garages or in space. For example, NASA space agency uses visual odometry in Mars Exploration Rovers missions for precise rover navigation on Martian terrain [4], [5].

We consider a specific setup where a stereo camera system is mounted on top of the car in the forward driving direction. Our goal is to develop a testbed for assessing the accuracy of various visual odometry implementations, which shall further be employed in higher level modules such as lane detection, lane departure or traffic sign detection and recognition. We decided to acquire our own GPS-registered stereo vision dataset since, to our best knowledge, none of the existing freely available datasets [6], [7] features a stereo-rig with inter-camera distance less than 20 cm (this distance is usually termed baseline). Additionally, we would like to be able to evaluate the impact of our camera calibration to the accuracy of the obtained results. Thus in this work we present a novel GPS-registered dataset acquired with a small-baseline stereo-rig (12 cm), the setup employed for its acquisition, as well as the results of some preliminary research.

In comparison with other similar work in this field [7], [8], we rely on low budget equipment for data acquisition. The groundtruth motion for our dataset has been acquired by a consumer grade GPS receiver. The employed GPS device doesnt have an inertial measurement unit, which means that we do not have access to groundtruth rotation and instead record only the WGS84 position in discrete time units. Additionally, hardware synchronization of the two sensors can not be performed due to limitations of the GPS device. Because of that, the alignment of the camera coordinates with respect to the WGS84 coordinates becomes harder to recover as will be explained later in the article. Thus our acquisition setup is much more easily assembled at the expense of more post-processing effort. However we shall see that the attained groundtruth accuracy is quite enough for drawing useful conclusions about several implementation details of the visual odometry.

## II. SENSOR SETUP

Our sensor setup consists of a stereo rig and a GPS receiver. The stereo rig has been mounted on top of the car, as shown in Fig. 1. The stereo rig (PointGrey Bumblebee2) features a Sony ICX424 sensor (1/3"), 12 cm baseline and global shutter. It is able to acquire two greyscale images 640×480 pixels each. The shutters of the two cameras are synchronized, which means that the two images of the stereo pair are acquired during the exactly same time interval (this is very important for stereo analysis of dynamic scenes). Both images of the stereo pair are transferred over one IEEE 1394A (FireWire) connection. The firewire connector is plugged into a PC express card connected to a laptop computer. The camera requires 12V power over the firewire cable. The PC express card is unable to draw enough power from the notebook and therefore features an external 12V power connector which we attach to the cigarette lighter power plug by a custom cable. In order to avoid overloading of the laptop bus, we set the frame rate of the camera to 25 Hz. The acquired stereo pairs are in the form of 640×480 pairs of interleaved pixels (16 bit), which means that upon acquisition the images need to be detached by placing each odd byte into the left image and each even byte into the right image. The camera firmware places timestamps in first four pixels of each stereo frame. These timestamps contain the value of a highly accurate internal counter at the time when the camera shutter was closed.

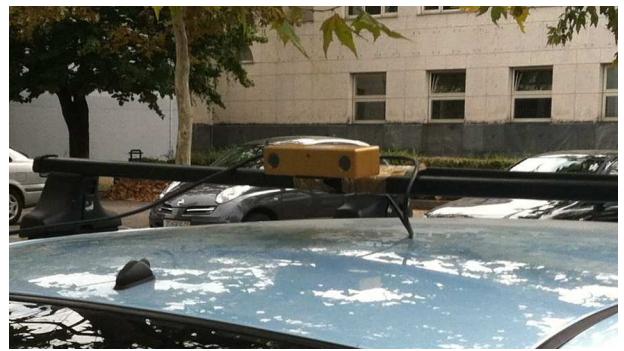

Fig. 1. The stereo system Bumblebee2 mounted on the car roof.

The employed GPS receiver (GeoChron SD logger) delivers location readings at 1 Hz. It is a consumer-grade device with a basic capability for multipath detection and mitigation. The GPS antenna was mounted on the car roof in close proximity to the camera. The GPS coordinates are converted from WGS84 to Cartesian coordinates in meters by a simple spherical transformation, which is sufficiently accurate given the size of the covered area. The camera and GPS are not mutually synchronized, so that the offset between camera time and GPS time has to be recovered in the postprocessing phase.

### III. DATASET ACQUISITION

The dataset has been recorded along a circular path throughout the urban road network in the city of Zagreb, Croatia. The path length was 688 m, the path duration was 111.4 s, while the top speed of the car was about 50 km/h. The dataset consists of 111 GPS readings and 2786 recorded frames with timestamps. The scenery was not completely static, since the video contains occasional moving cars in both directions, as well as pedestrians and cyclists.

The acquisition was conducted at the time of day with the largest number of theoretically visible satellites (14). For comparison, the least value of that number at that day was 8. In practice, our receiver established connection to 9.5 satellites along the track, on average. Thus, at 99.1% locations we had HDOP (horizontal dilution of precision) below 1.3 m while HDOP was less than 0,9 m at 57.1% locations.

The obtained GPS accuracy has been qualitatively evaluated by plotting the recorded track over a rectified digital aerial image of the site, as shown in Fig.2. The figure shows that the GPS track follows the right side of the road accurately and consistently, except at bottom right where our car had to avoid parked cars and pedestrians. Thus, the global accuracy of the recorded GPS track appears to be in agreement with the HDOP figures stated above. Furthermore, the relative motion between the neighbouring points which we use in our quantitative experiments is much more accurate than that and approaches

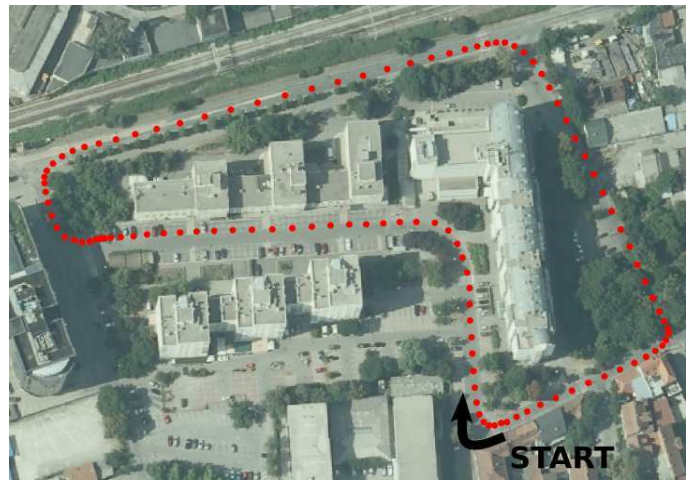

Fig. 2. Projection of the acquired GPS track onto rectified aerial imagery.

DGPS accuracy. We note that the aerial orthophoto from the figure has been provided by Croatian Geodetic Administration, while the conversion of WGS84 coordinates into Croatian projected coordinates (HTRS96) and visualization are carried out in Quantum GIS. For additional verification, the same experiment has been performed in Google Earth as well, where we obtained very similar results.

### IV. VISUAL ODOMETRY

As stated in the introduction, the visual odometry recovers incremental motion between subsequent images of a video sequence by exclusively relying on image correspondences. We shall briefly review the main steps of the technique in the case of a calibrated and rectified stereo system. Calibration of the stereo system consists of recovering internal parameters of the two cameras such as the field of view, as well as exact position and orientation of the second camera with respect to the first one. Rectification consists in transforming the two acquired images so they correspond to images which would be obtained by a system in which the viewing directions of the two cameras are mutually parallel and orthogonal to the baseline. Calibration and rectification significantly simplify the procedure of recovering inter-frame-motion as we outline in the following text.

One typically starts by finding and matching the corresponding features in the two images. Because our system is rectified the search for correspondences can be limited to the same row of the other image. Since the stereo system is calibrated, these correspondences can be easily triangulated and subsequently expressed in metric 3D coordinates. Then the established correspondences are tracked throughout the subsequent images. In order to save time, tracking and stereo matching typically use lightweight point features such as corners or blobs [9]–[11]. The new positions of previously triangulated points provide constraints for recovering the new location of the camera system. The new location can be recovered either analytically, by recovering pose from projections of known 3D points [12], [13], or by optimization [11], [14], [15]. The location estimation is usually performed only in some images of the acquired video sequence, which are often referred to as key-images.

Thus, we have seen that the visual odometry is able to provide the camera motion between the neighboring key-images. This motion has 6 degrees of freedom (3 rotation and 3 translation), and we represent this motion by a 4×4 transformation matrix which we denote by $[\mathbf{Rt}]$. This matrix contains parameter of the rotation matrix $\mathbf{R}$ and the translation vector $\mathbf{t}$, as shown in equation (1).

$$[\mathbf{Rt}] = \begin{bmatrix} r_{11} & r_{12} & r_{13} & t_1 \\ r_{21} & r_{22} & r_{23} & t_2 \\ r_{31} & r_{32} & r_{33} & t_3 \\ 0 & 0 & 0 & 1 \end{bmatrix} \quad (1)$$

From a series of $[\mathbf{Rt}]$ matrices we can calculate the full path trajectory by cumulative matrix multiplication. This is shown in equation (2) where the camera location $\mathbf{p}_{vo}(t_{\text{CAM}})$ is determined by multiplying matrices from the key-image 1 to the key-image $t$.

$$\mathbf{p}_{vo}(t_{\text{CAM}}) = \prod_{i=1}^{t} [\mathbf{Rt}]_i \quad (2)$$

At this moment we must observe that the recovered camera locations $\mathbf{p}_{vo}(t_{\text{CAM}})$ are expressed in the temporal and spatial coordinates of the 0th key-image which is therefore acquired at time $t_{\text{CAM}} = 0\,\text{s}$. The coordinate axes $x$ and $y$ of the trajectory $\mathbf{p}_{vo}(t_{\text{CAM}})$ are aligned with the corresponding axes of the 0th key-image, while the z axis is perpendicular to that image. The locations estimated by visual odometry will be represented in this coordinate system by accumulating all estimated transformations from the 0th frame. If we wish to relate these locations with GPS readings, we shall need to somehow recover the translation and rotation of the 0th key-frame with respect to the GPS coordinates. We shall achieve that by aligning the first few incremental motions with the corresponding GPS motions. However, before we do that, we first need to achieve the temporal synchronization between the two sensors.

## V. Sensor synchronization

As stated in section II, the camera and GPS receiver are not synchronized. Thus, we need to recover the time interval $t_{vo}^0$ between the 0th video frame and the 0th GPS location, such that the camera time $t_{\text{CAM}}$ corresponds to $t + t_{vo}^0$ in GPS time ($t = 0$ corresponds to the 0th GPS location). We estimate $t_{vo}^0$ by comparing absolute incremental displacements of trajectories $\mathbf{p}_{gps}(t)$ and $\mathbf{p}_{vo}(t_{\text{CAM}})$ obtained by GPS and visual odometry, respectively. In order to do that, we first define $\Delta\mathbf{p}(t, \Delta t)$ as the incremental translation at time $t$:

$$\Delta\mathbf{p}(t, \Delta t) = \mathbf{p}(t) - \mathbf{p}(t - \Delta t) \,. \quad (3)$$

We also define $\Delta s(t)$ as the absolute travelled distance during the previous interval of $\Delta t_{\text{GPS}}=1\,\text{s}$ (GPS frequency is 1 Hz):

$$\Delta s(t) = \|\Delta\mathbf{p}(t, \Delta t_{\text{GPS}})\|, t \in \mathbb{N} \,. \quad (4)$$

Now, if we consider the time instants in which the GPS positions are defined (that is, integral time in seconds), we can pose the following optimization problem:

$$\hat{t}_{vo}^0 = \underset{t_{vo}^0}{\operatorname{argmin}} \sum_{t=1}^{T_{\text{last}}-1} (\Delta s_{vo}(t + t_{vo}^0) - \Delta s_{gps}(t))^2 \,. \quad (5)$$

The problem is well-posed since the absolute incremental displacements $\Delta s_{vo}$ and $\Delta s_{GPS}$ are agnostic with respect to the fact that the camera and GPS coordinate systems are still misaligned. We see that in order to solve this problem by optimization, we need to interpolate all locations obtained by visual odometry at times of GPS points for each considered time offset $t_{vo}^0$. However, that does not pose a computational problem due to very low frequency of the GPS readings (there are only 111 GPS locations in our dataset). Thus the problem can be easily solved by any optimization algorithm, and so we can express visual odometry locations in GPS time $\mathbf{p}_{vo'}$ as:

$$\mathbf{p}_{vo'}(t) = \mathbf{p}_{vo}(t + t_{vo}^0) \,. \quad (6)$$

The interpolation is needed due to the fact that we capture images at approximately 25 Hz and GPS data with 1 Hz (cf. section II). The time intervals between two subsequent images often differ from expected 40 ms due to unpredictable bus congestions within the laptop computer (this is the reason why camera records timestamps in the first place). We recover the locations of visual odometry "in-between" the acquired frames by the following procedure. We first accumulate timestamps in the frame sequence until we reach the two frames which are temporally closest to the desired GPS time. Finally we determine the desired location between these two frames using linear interpolation.

We also can relate the two trajectories by the absolute incremental rotation angle $\Delta\phi$ between the corresponding two time instants. We can recover this angle by looking at three consecutive locations as follows:

$$\Delta\phi(t) = \arccos \frac{\langle \Delta\mathbf{p}(t, \Delta t_{\text{GPS}}), \Delta\mathbf{p}(t+1, \Delta t_{\text{GPS}}),\rangle}{\|\Delta\mathbf{p}(t, \Delta t_{\text{GPS}})\|\|\Delta\mathbf{p}(t+1, \Delta t_{\text{GPS}})\|} \quad (7)$$

This procedure is illustrated in Fig. 3. Thus we propose two metrics suitable for relating the trajectories obtained by GPS and visual odometry: $\Delta s(t)$ and $\Delta\phi(t)$. Note that in order to be able to determine these metrics for visual odometry at GPS times, one needs to apply the previously recovered offset $t_{vo}^0$ and employ interpolation between the closest image frames.

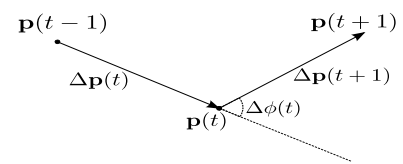

Fig. 3. The absolute incremental rotation angle $\Delta\phi$ can be determined from incremental translation vectors.

## VI. Aligning the sensor coordinates

Now that we have synchronized data between camera and GPS, we need to estimate the 3D alignment of the reference coordinate system of the visual odometry with respect to the GPS coordinate system. In other words we need to find a rigid transformation between the 0th GPS location and the location of the 0th camera frame. After this is completed, we shall be able to illustrate the overall accuracy of the visual odometry results with respect to the GPS readings.

The translation between the two coordinate systems can be simply expressed as:

$$\mathbf{T}_{\text{vo}}^{\text{GPS}} = \mathbf{p}_{GPS}(0\,s) - \mathbf{p}_{vo'}(0\,s)\,. \qquad (8)$$

In order to recover rotation alignment, we consider incremental translation vectors for visual odometry ($\Delta \mathbf{p}_{vo'}(t)$) and GPS ($\Delta \mathbf{p}_{gps}(t)$) between two consecutive GPS times, as determined in (3). We find the optimal rotation alignment of the visual odometry coordinate system $\mathbf{R}_{\text{vo}}^{\text{GPS}}$ by minimizing the following error function:

$$\hat{\mathbf{R}}_{\text{vo}}^{\text{GPS}} = \underset{\mathbf{R}}{\arg\min} \sum_{t=1}^{N} \left[ \arccos \frac{\langle \mathbf{R} \cdot \Delta \mathbf{p}_{vo'}(t), \Delta \mathbf{p}_{gps}(t) \rangle}{\|\Delta \mathbf{p}_{vo'}(t)\| \|\Delta \mathbf{p}_{gps}(t)\|} \right]^2 \qquad (9)$$

In order to bypass the accumulated noise which grows over time, we choose $N = 4$. This problem is easily solved by any nonlinear optimization algorithm, and so we can express visual odometry locations in GPS Cartesian coordinates and GPS time $\mathbf{p}_{vo''}$ as:

$$\mathbf{p}_{vo''}(t) = \mathbf{R}_{\text{vo}}^{\text{GPS}} \cdot \mathbf{p}_{vo'}(t) + \mathbf{T}_{\text{vo}}^{\text{GPS}}\,. \qquad (10)$$

We note that this approach would be underdetermined if we had only straight car motion at the beginning of video, and to avoid that we arranged that our dataset begins on the road turn. Experiments showed that this approach works well in practice. Better accuracy could be obtained by employing a GPS sensor capable of producing rotational readings, however that would be out of the scope of this work.

## VII. EXPERIMENTS AND RESULTS

In our experiments we employ the library Libviso2 which recovers the 6 DOF motion of a moving stereo rig by minimizing reprojection error of sparse feature matches [11]. Libviso2 requires rectified images on input. Currently, we consider each third frame from the dataset because, with 25 fps, camera movement between two consecutive frames is very small with urban driving speeds and such small camera movements often add more noise than useful data. We calibrated our stereo rig on two different calibration datasets by means of the OpenCV module calib3d. In the first case the calibration pattern was printed on A4 paper, while in the second the calibration pattern was displayed on 21" LCD screen with HD resolution (1920x1080). In both cases, the raw calibration parameters have been employed to rectify the input images and to produce rectified calibration parameters which are supplied to Libviso2. The resulting dataset and calibration parameters is freely downloadable for research purposes from http://www.zemris.fer.hr/~ssegvic/datasets/fer-bumble.tar.gz.

### A. A4 calibration

We first present the results obtained with the A4 calibration dataset. The resulting trajectories $\mathbf{p}_{vo''}(t)$ and $\mathbf{p}_{gps}(t)$ are compared in Fig.4. We note that the shape of the trajectory is mostly preserved, however there is a large deviation in scale. We compare the corresponding incremental absolute displacements $\Delta s_{gps}(t)$ and $\Delta s_{vo''}(t)$, and the scale error $\Delta s_{gps}(t)/\Delta s_{vo''}(t)$ In Fig.5. We observe that the two graphs are well aligned, and that the visual odometry generally undershoots in translation motion compared to the GPS groundtruth.

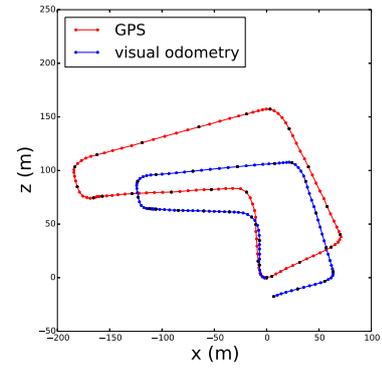

Fig. 4. Comparison between the GPS trajectory and the recovered visual odometry trajectory with the A4 calibration.

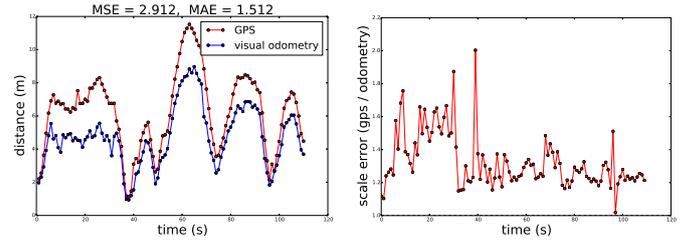

Fig. 5. Comparison of incremental translations (left) and the resulting scale error (right) as obtained with GPS and visual odometry with the A4 calibration.

Note that the scale error is not constant. Thus the results could not be improved by simply fixing the baseline recovered by the calibration procedure.

### B. LCD calibration

We now present the results obtained with the LCD calibration dataset. The GPS trajectory $\mathbf{p}_{gps}(t)$ and the resulting visual odometry trajectory $\mathbf{p}_{vo''}(t)$ are shown in Fig.6. By comparing this figure with Fig.4, we see that a larger calibration pattern produced a huge impact on final results.

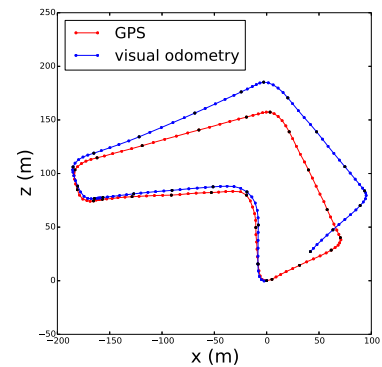

Fig. 6. Comparison between the GPS trajectory and the recovered visual odometry trajectory with the LCD calibration.

### C. Influence of distant features

While analyzing Libviso2 source code, we noticed that it rounds all zero disparities to $1.0$. This is necessary since otherwise there would be a division by zero in the triangulation procedure. However, this means that all features with zero disparity are triangulated on a plane which is much closer to the camera than the infinity where it should be (note that the distance of that plane depends on the baseline). In order to investigate the influence of that decision to the recovered trajectory, we changed the magic number from 1.0 to 0.01. The effects of that change are shown in Fig.7, where we note a significant improvement with respect to Fig.6.

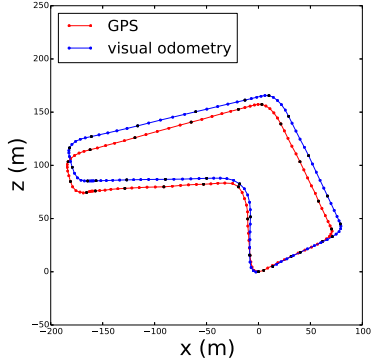

Fig. 7. Comparison between the GPS trajectory and the recovered visual odometry trajectory with the LCD calibration and library correction.

The absolute incremental translations and the resulting scale errors for this case are shown in Fig.8. The corresponding absolute incremental rotations are shown in Fig.9. We note a very nice alignment for incremental translation, and somewhat less successful alignment for incremental rotation. Note that large discrepancies in absolute incremental rotation at times 35 s and 97 s occur at low speeds (cf. Fig.8 (left)), when the equation (7) becomes less well-conditioned.

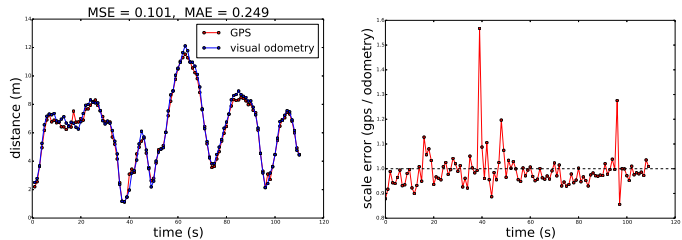

Fig. 8. Comparison of incremental translations (left) and the resulting scale error (right) as obtained with GPS and visual odometry with the LCD calibration and library correction.

Note that this issue could also have been solved by simply neglecting features with zero disparity. However, that would be wasteful since the features at infinity provide valuable constraints for recovering the rotational part of inter-frame motion. We believe that this has been overlooked in the original library since Libviso was originally tested on a stereo setup with 4x larger baseline and 2x larger resolution [11]. Thus the original setup entails much less features with zero disparity, while the effect of these features to the reconstruction accuracy is not easily noticeable.

### D. Quantitative comparison of the achieved accuracy

We assess the overall achieved accuracy of the recovered trajectory $\mathbf{p}_{vo''}(t)$ by relying on previously introduced incremental metrics $\Delta\phi$ and $\Delta s$. These metrics shall now be used to define mean square error ($MSE$) and mean absolute errors ($MAE$) for quantitative assessment of the achieved accuracy.

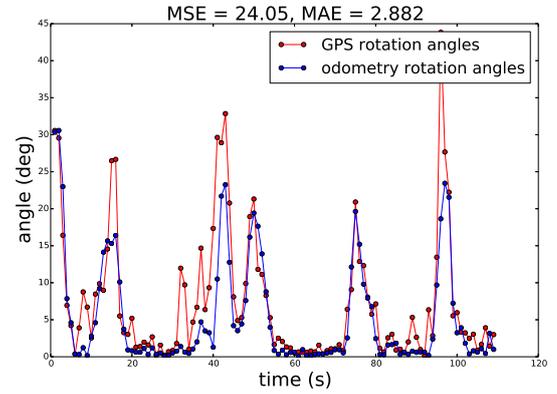

Fig. 9. Comparison of incremental rotations as obtained with GPS and visual odometry with the LCD calibration and library correction.

$$MSE_{trans} = \frac{1}{N}\sum_{t=1}^{N}(\Delta s_{vo''}(t) - \Delta s_{gps}(t))^2 \quad (11)$$

$$MAE_{trans} = \frac{1}{N}\sum_{t=1}^{N}|\Delta s_{vo''}(t) - \Delta s_{gps}(t)| \quad (12)$$

$$MSE_{rot} = \frac{1}{N}\sum_{t=1}^{N}(\Delta\phi_{vo''}(t) - \Delta\phi_{gps}(t))^2 \quad (13)$$

$$MAE_{rot} = \frac{1}{N}\sum_{t=1}^{N}|\Delta\phi_{vo''}(t) - \Delta\phi_{gps}(t)| \quad (14)$$

The obtained results are presented in Table I. The rows of the table correspond to the metrics $MSE_{\text{rot}}$, $MAE_{\text{rot}}$, $MSE_{\text{trans}}$ and $MAE_{\text{trans}}$. The columns correspond to the original library with the A4 calibration (A4), the original library with the LCD calibration (LCD1), and the corrected library with the LCD calibration (LCD2). A considerable improvement is observed between A4 and LCD1, while the difference between LCD1 and LCD2 is still significant.

TABLE I.  MSE AND MAE ERRORS IN DEPENDENCE OF CAMERA CALIBRATION AND DISTANT FEATURES TRIANGULATION.

| cases: | A4 | LCD1 | LCD2 |
|---|---|---|---|
| $MSE_{\text{trans}}(m^2)$ | 2.912 | 0.111 | 0.101 |
| $MAE_{\text{trans}}(m)$ | 1.512 | 0.260 | 0.249 |
| $MSE_{\text{rot}}(deg^2)$ | 24.914 | 24.856 | 24.053 |
| $MAE_{\text{rot}}(deg)$ | 2.967 | 2.897 | 2.882 |

### E. Experiments on the artificial dataset

The previous results show that, alongside camera calibration, the feature triangulation accuracy has a large impact on the accuracy of visual odometry. In this section we explore that finding on different baselines of the stereo rig. To do this, we develop an artificial model of a rectified stereo camera system. On input, the model takes camera intrinsic and extrinsic parameters as well as some parameters of the camera motion. Then it generates an artificial 3D point cloud and projects it to the image plane in every frame of camera motion. The point cloud is generated in a way so that its projections resemble features we encounter in real world while analysing imagery acquired from a driving car. On output, the model produces the feature tracks which are supplied as input to the library for visual odometry as input. Fig.10 shows the comparison of the straight trajectory in relation to different camera baseline setups. As one would expect, the obtained accuracy significantly drops as the camera baseline is decreased. Furthermore, Fig.11 shows the results on the same dataset after modification in treatment of zero-disparity features inside the library. The graphs in the figure show that for small baselines the modified library performs increasingly better than the original. This improvement occurs since the number of zero-disparity features increases as the baseline becomes smaller.

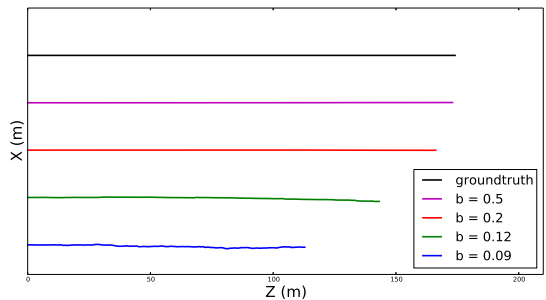

Fig. 10. Reconstructed trajectories with original Libviso2.

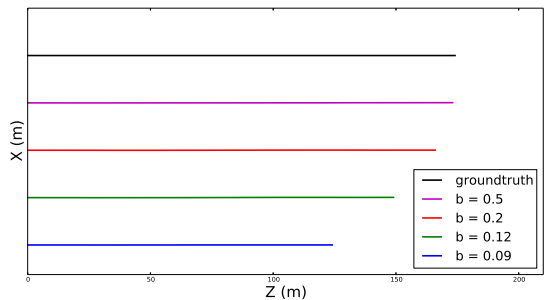

Fig. 11. Reconstructed trajectories with modified Libviso2.

## VIII. Conclusion and future work

We have proposed a testbed for assessing the accuracy of visual odometry with respect to the readings of an unsynchronized consumer-grade GPS sensor. The testbed allowed us to acquire an experimental GPS-registered dataset suitable for evaluating existing implementations in combination with our own stereo system. The acquired dataset has been employed to assess the influence of calibration dataset and some implementation details to the accuracy of the reconstructed trajectories.

The obtained experimental results show that a consumer grade GPS system is able to provide useful groundtruth for assessing performance of visual odometry. This still holds even if the synchronization and alignment with the camera system is performed at the postprocessing stage. The experiments also show that the size and the quality of the calibration target may significantly affect the reconstruction accuracy. Additionally, we have seen that features with zero disparity should be treated with care, especially in small-baseline setups. Finally, the most important conclusion is that small-baseline stereo systems can be employed as useful tools in SfM analysis of video from the driver's perspective.

Our future research shall be directed towards exploring the influence of other implementation details to the reconstruction accuracy. The developed implementations shall be employed as a tool for improving the performance of several computer vision applications for driver assistance, including lane recognition, lane departure warning and traffic sign recognition. We also plan to collect a larger dataset corpus which would contain georeferenced videos acquired by stereo rigs with different geometries.

### Acknowledgment

This work has been supported by research projects Vista (EuropeAid/131920/M/ACT/HR) and Research Centre for Advanced Cooperative Systems (EU FP7 #285939).